\documentclass{article}

\usepackage[preprint]{neurips_2023}

\usepackage[utf8]{inputenc} 
\usepackage[T1]{fontenc}    
\usepackage{hyperref}       
\usepackage{url}            
\usepackage{booktabs}       
\usepackage{amsfonts}       
\usepackage{nicefrac}       
\usepackage{microtype}      
\usepackage{xcolor}         

\usepackage{amsmath}
\usepackage{graphicx}
\usepackage{changepage}
\usepackage{tabu}
\usepackage{rotating}

\usepackage{textcmds}
\usepackage{subcaption}
\usepackage{multirow}
\usepackage{bm}

\usepackage{tikz}
\usepackage{amsmath,bm,times}
\usetikzlibrary{fadings}
\usetikzlibrary{shapes.arrows}
\usetikzlibrary{calc}

\usepackage{sistyle}
\SIthousandsep{,}

\title{Kindness in Multi-Agent Reinforcement Learning}

\author{		
	Farinaz Alamiyan-Harandi \\
	Department of Electrical \& Computer Engineering \\
	Isfahan University of Technology \\
	Isfahan 84156-83111, Iran \\
	\texttt{farinaz.alamiyan@gmail.com} \\ 
	\And
	Mersad Hassanjani \\
	Department of Electrical \& Computer Engineering \\
	Isfahan University of Technology \\
	Isfahan 84156-83111, Iran \\
	\texttt{mersadhassanjani@gmail.com} \\ 	
	\And
	Pouria Ramazi \\
	Department of Mathematics \& Statistics \\ Brock University\\
	St. Catharines, ON L2S 3A1, Canada\\  
	\texttt{p.ramazi@gmail.com} \\
}

\begin{document}
	
	\pgfdeclarelayer{background}
	\pgfdeclarelayer{foreground}
	\pgfsetlayers{background,main,foreground}
	
	\tikzstyle{states}=[dotted, black, fill=green!40, text width=1.2em, text centered, minimum height=1.7em, thick]
	
	\tikzstyle{layers} = [draw, thick, text centered, text width=2.5em, fill=white!20, 
	minimum height=6em]
	\def\blockdist{0.5}
	\def\edgedist{1}
	\definecolor{dodgerblue}{rgb}{0.12, 0.56, 1.0}

\maketitle

\begin{abstract}
  In human societies, people often incorporate fairness in their decisions and treat reciprocally by being kind to those who act kindly. They evaluate the kindness of others' actions not only by monitoring the outcomes but also by considering the intentions. This behavioral concept can be adapted to train cooperative agents in Multi-Agent Reinforcement Learning (MARL). We propose the KindMARL method, where agents' intentions are measured by counterfactual reasoning over the environmental impact of the actions that were available to the agents. More specifically, the current environment state is compared with the estimation of the current environment state provided that the agent had chosen another action. The difference between each agent's reward, as the outcome of its action, with that of its fellow, multiplied by the intention of the fellow is then taken as the fellow's \qq{kindness}. If the result of each reward-comparison confirms the agent's superiority, it perceives the fellow's kindness and reduces its own reward. Experimental results in the Cleanup and Harvest environments show that training based on the KindMARL method enabled the agents to earn $89\%$ (resp. $37\%$) and $44\%$ (resp. $43\%$) more total rewards than training based on the Inequity Aversion and Social Influence methods. The effectiveness of KindMARL is further supported by experiments in a traffic light control problem. 
\end{abstract}

\section{Introduction}
\label{S.Introduction}
Reinforcement Learning (RL) is a promising framework of machine learning where an intelligent agent interacts within a given environment, by applying \textit{action}s and receiving \textit{reward}s, to learn an optimal \textit{policy}, a probability distribution over the available actions that maximizes the total obtained rewards for each visited environment \textit{state} \citep{sutton1998introduction}. 
RL tackles a wide range of tasks by using a simple concept of \textit{rewards and punishments} without requiring prior knowledge about the given problem. 
Many real-world problems such as resource allocation \citep{naderializadeh2021resource} and real-time traffic flow control \citep{wei2019colight} consider group activities in an environment with shared resources where a conflict of interest exists between individual and social preferences. 
To handle these problems, the single-agent RL algorithms are extended as Multi-Agent RL (MARL) settings, where each agent learns a policy from its own observations. 
However, partial observability and the environmental impact of the actions applied by other agents raise uncertainty in the environmental states and makes the training process unstable \citep{chu2019multi}. 
Learning communication protocols to exchange the agents' information \citep{foerster2016learning} and enriching the \textit{extrinsic reward}s received from the environment by using some \textit{intrinsic reward}s \citep{jaques2019social} are techniques that are introduced to stabilize training and improve the learned policies in MARL approaches.

Intrinsic rewards encode some aspects of agents' behavior that are not represented by extrinsic rewards explicitly. 
They are scalar signals that an agent internally computes and are typically defined based on the behavioral practices and social concepts in human societies. 
Examples of these concepts incorporated in MARL as intrinsic rewards are as follows: 
\emph{(i) social influence}, the agent's ability to select the action that influences its fellows' next action \citep{jaques2019social},
\emph{(ii) joint attention}, the agent's ability to manage its attention to the same object or event that one of its fellows attends \citep{lee2021joint}, 
\emph{(iii) curiosity}, the agent's capability to predict the outcomes of its own actions \citep{heemskerk2020social}, 
\emph{(iv) environmental impact}, the agent's criterion to measure the fellows' role in reaching the current rewarding environmental state \citep{EMuReL2023}, and 
\emph{(v) inequity aversion}, the agent's comparison-based criterion to judge its fellows' cooperation \citep{hughes2018inequity}. 

Consider the Inequity Aversion (IA) method \citep{hughes2018inequity}, for example, where each agent changes its own extrinsic reward based on the envy and guilt concepts by comparing its own extrinsic reward with that of each of its fellows. If the result of a pair-comparison confirms the agent's \textit{advantageous inequity}, that it has earned more than its fellow, then the agent feels guilty and if the result confirms the agent's \textit{disadvantageous inequity}, the agent experiences envy. In both situations, the agent reduces its own reward as a punishment.

We aim to enrich the IA model based on the theory of reciprocity \citep{falk2006theory} in social psychology, stating that consequences of an action are not the only criterion to evaluate its kindness--the actor's intentions also influence others' conclusions.
In MARL, an agent's intention can be measured by its impact in reaching the current environment state that resulted in the agents' rewards \citep{EMuReL2023}. 
To this end, we propose the KindMARL method. 
We equip each agent with the Extended Intrinsic Curiosity Module (EICM) \citep{EMuReL2023} to predict the role of every other agent's actions in reaching the current environment state. 
An agent considers the action set of its fellow as the fellow's alternatives and predicts the follow's counterfactual roles in achieving the current environment state. 
Then the agent measures its fellow's intention by dividing the impact of the fellow's action to the maximum impact obtained by any of the fellow's available actions. 
Multiplying every IA pair-comparison by the intention of its corresponding fellow, the agent becomes sensitive to others' intentions as well as their rewards and reacts to other agents differently according to their kindness. 
The agent, hence, can act fairer in dealing with itself as well as the others. 
We test the KindMARL method on CLeanup and Harvest environments (the \emph{common pool resource problems} \citep{leibo2017multi}) and also the problem of traffic light control. We compare the results with those of the state-of-the-art methods.

The remaining of this paper is organized as follows: 
Section \ref{S.MARL_method} formalizes the MARL problem and describes the proposed KindMARL method.
The case study and experimental results are given in Section \ref{S.Experiments} and are discussed in Section \ref{S.Discussion}.


\section{The KindMARL approach}
\label{S.MARL_method}
We consider the following MARL problem: $N$ agents coexist in an environment with the state space $\mathcal{S}$. 
At each time step $t$, each agent $k$ partially observes the environment's \textit {global state} $s_t$ as the \textit{local state} $s^k_t$, and selects an action $a^k_t$ from the action set $\mathcal{A}^k=\{a_1,...,a_m\}$ by using a \textit{policy} $\pi^k$, that is a probability distribution over the agent's action set $\mathcal{A}^k$. 
By applying the joint action $\bm{a_t}=[a^1_t,...,a^k_t,...,a^N_t]$ at the global state $s_t$, the environment transfers to the global state $s_{t+1}$ according to a transition distribution $\mathcal{T}(s_{t+1}|s_{t},\bm{a}_{t})$, resulting the reward $r^k_{t+1}$ to each agent $k$.
The reward is used it to compute a \textit{state-action value function} $Q^{\pi^k}(s^k_t,a^k_t)$ for all local states $s^k_t$ and agent $k$'s action set $\mathcal{A}^k$, where $Q^{\pi^{k}}(s^{k}_{t},a^k_t)$ approximates the \textit{return} $R^{k}_{t}=\sum_{i=0}^{\infty}\gamma^{i}r^{k}_{t+i+1}$, that is an estimation of the cumulative $\gamma$- discounted rewards over all local states visited in future by applying action $a^k_t$ on local state $s^k_t$ and following policy $\pi^k$. Agent $k$ selects the action with maximum $Q^{\pi^k}(s^k_t,a^k_t)$ in each time step $t$. 

To train cooperative agents, we use the linear reward function $r^k_t = \alpha e^k_t + \beta i^k_t $ where $\alpha$ and $\beta$ are constant scalars, $e^k_t$ is the extrinsic reward that agent $k$ receives from the environment, and $i^k_t$ is the intrinsic reward that agent $k$ computes by incorporating its fellows' kindness in the IA model \citep{hughes2018inequity} as follows:
\begin{equation}\label{equ.impact_Inequity_Aversion_reward}
\begin{aligned}
i_{t}^{k} = -\frac{\alpha_{k}}{N-1}\sum_{j\neq k} d^{k,j}_{t}\max(w^{j}_{t} - w^{k}_{t}, 0) - \frac{\beta_{k}}{N-1}\sum_{j\neq k}  d^{k,j}_{t}\max(w^{k}_{t} - w^{j}_{t}, 0),
\end{aligned}
\end{equation}
where parameters $\alpha_{k}$,$\beta_{k}\in\mathbb{R}$ are adjustable parameters of the IA model, $w^j_t$ as a temporary memory of the extrinsic reward occurrence \citep{hughes2018inequity} is computed as 
\begin{equation}\label{equ.Inequity_Aversion_e}
\begin{aligned}
w^{j}_{t} = \gamma \lambda w^{j}_{t-1} + e_{t}^{j}  \;\;\;\;\;\;\;\forall t\geq 1, \; w^j_0 = 0,
\end{aligned}
\end{equation}
where $\lambda\in[0,1]$ is a trace-decay hyper-parameter, and $d^{k,j}_{t}$ is agent $j$'s \textit{intention} computed by agent $k$.

To measure the fellows' intention as the main factor of evaluating kindness, agent $k$ estimates the environment impact of each of its fellows considering counterfactual reasoning. 
It estimates what would happen in the current state if agent $j$ had acted differently. 
In other words, agent $k$ predicts its current local state $s^k_t$ by supposing each agent $j$ applies one of its other available actions. 
To this end, agent $k$ uses the forward model of EICM structure \citep{EMuReL2023}, that estimates the current encoded local state of the environment ($\hat{\phi}(\phi(s^k_{t-1}),\bm{a}_{t-1})$) by receiving the previous encoded local state $\phi(s^k_{t-1})$ and the previous joint action, $\bm{a}_{t-1}$, where $\phi$ is an encoding function. 
Agent $k$ estimates the current encoded local state by replacing agent $j$'s action with one of its counterfactual actions ${\tilde{a}^{j}_{t-1}}=\{b | b\neq a^{j}_{t-1},b\in A\}$. 
Then it measures the difference between the previous encoded local state $\phi(s^{k}_{t-1})$ with the counterfactual prediction of the current encoded local state $\hat{\phi}_{\tilde{a}^{j}_{t-1}}(\phi(s^k_{t-1}),\bm{a}^{\tilde{j}}_{t-1})$ by using the mean square error function as follows:
\begin{equation}\label{equ.Lt}
\begin{aligned}
{L_{t}}= \frac{1}{2} \left \| \phi(s^k_{t-1}) - \hat{\phi}(\phi(s^k_{t-1}),\bm{a}^{\tilde{j}}_{t-1}))\right \| _{2}^{2}, 
\end{aligned}
\end{equation}
where $\bm{a}^{\tilde{j}}_{t-1}$ indicates that agent $j$'s action in the joint action $\bm{a}_{t-1}$ is replaced by one of its counterfactual actions. 
Let $l$ be the difference that is computed by using the joint action $\bm{a_{t-1}}$ and ${L}^{j}_{t}$ be the list of differences, that each of its elements is calculated considering one counterfactual action of agent $j$. 
Then agent $j$'s intention is computed as follows (Figure. \ref{fig.EICM_counterfactual}):
\begin{equation}\label{equ.Deservedness_Counterfactual}
\begin{aligned}
d^{k,j}_t= \frac{l}{\max(L^j_t)} .
\end{aligned}
\end{equation}
\begin{center}
	\begin{figure}[!h]
		\centering
		\resizebox{\columnwidth}{!}{
			\begin{tikzpicture} 
			\node (Input11) at (5.9,0.5) [] {$\phi(s^k_{t-1})$};
			
			\node (Input21) at (-0.2,-0.45) [] {$\phi(s^k_{t-1})$};
			\node (Input22) at (0,-1.3) [] {$u^k_{t-1}$};
			\node (Input23) at (-0.3,-1.9) [fill=magenta!35] {$\bm{a}^{\tilde{j}}_{t-1}$};			
			\node (Input231) at (-0.2,-2) [fill=magenta!25] {$\bm{a}^{\tilde{j}}_{t-1}$};
			\node (Input232) at (-0.1,-2.1) [fill=magenta!15] {$\bm{a}^{\tilde{j}}_{t-1}$};
			\node (Input233) at (0,-2.2) [fill=magenta!5] {$\bm{a}^{\tilde{j}}_{t-1}$};
			
			\node (ForwardModel) at (2,-1.3) [layers, text width=5em, rectangle, draw=magenta!100, minimum height=6.5em, text=magenta!100, thick] {Forward Model};
			
			\node (Output) at (5,-1) [fill=magenta!35] {$\hat{\phi}(\phi(s^k_{t-1}),\bm{a}^{\tilde{j}}_{t-1})$};
			\node (Output) at (5.1,-1.1) [fill=magenta!25] {$\hat{\phi}(\phi(s^k_{t-1}),\bm{a}^{\tilde{j}}_{t-1})$};
			\node (Output) at (5.2,-1.2) [fill=magenta!15] {$\hat{\phi}(\phi(s^k_{t-1}),\bm{a}^{\tilde{j}}_{t-1})$};
			\node (Output) at (5.3,-1.3) [fill=magenta!5] {$\hat{\phi}(\phi(s^k_{t-1}),\bm{a}^{\tilde{j}}_{t-1})$};
			
			\draw [->, thick] (Input21.east) -- (ForwardModel.140);  
			\draw [->, thick] (Input22.east) -- (ForwardModel.180); 
			\draw [->, thick] (Input233.east) -- (ForwardModel.222); 
			\draw [->, thick] (ForwardModel.east) -- (Output.west); 
			\coordinate (Origin) at (5,-1.2);
			\draw [<->, draw=black!30!green!90, thick, dashed] ($(Origin)+(1.75,0)$) 
			arc (-35:35:1.5);
			
			\node at (10.5,-0.2) [fill=magenta!35] {$\frac{1}{2} \left \| \phi(s^k_{t-1}) - \hat{\phi}(\phi(s^k_{t-1}),\bm{a}^{\tilde{j}}_{t-1}))\right \| _{2}^{2}$};
			\node at (10.6,-0.3) [fill=magenta!25] {$\frac{1}{2} \left \| \phi(s^k_{t-1}) - \hat{\phi}(\phi(s^k_{t-1}),\bm{a}^{\tilde{j}}_{t-1}))\right \| _{2}^{2}$};
			\node at (10.7,-0.4) [fill=magenta!15] {$\frac{1}{2} \left \| \phi(s^k_{t-1}) - \hat{\phi}(\phi(s^k_{t-1}),\bm{a}^{\tilde{j}}_{t-1}))\right \| _{2}^{2}$};
			\node at (10.8,-0.5) [fill=magenta!5] {$\frac{1}{2} \left \| \phi(s^k_{t-1}) - \hat{\phi}(\phi(s^k_{t-1}),\bm{a}^{\tilde{j}}_{t-1}))\right \| _{2}^{2}$};
			
			\node at (7.5,-0.2) {$L^j_t$};
			\node at (10.8,-2) {$d^{k,j}_t= \frac{l}{\max(L^j_t)}$};				
			
			\node at (10.8,-1.2) [
			right color=pink!30!pink!50,
			single arrow,
			single arrow head extend=0.35cm,
			single arrow tip angle=110,
			single arrow head indent=0.1cm,
			minimum height=0.5cm,
			inner sep=4pt,
			shading angle=90+90,
			rotate=-90
			] {};
			\end{tikzpicture}}
		\caption{The process of computing agent $j$'s intention by agent $k$.} 
		\label{fig.EICM_counterfactual}
	\end{figure}
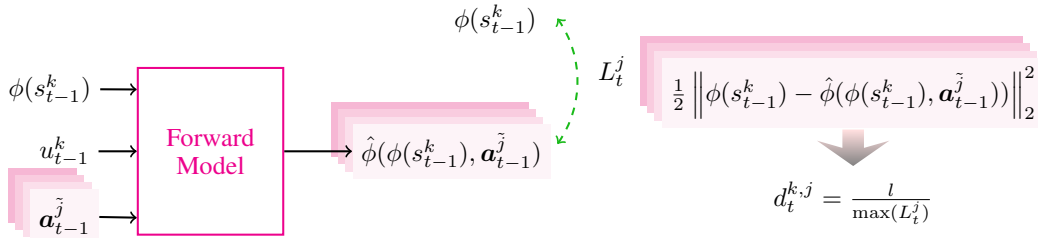
\end{center}
\section{Experiments}
\label{S.Experiments}
\subsection{The problem of Cleanup and Harvest}
Cleanup is a public goods game wherein agents must clean a river before apples can develop, but aren't able to harvest apples at the same time as cleaning. In Cleanup, agents should effectively time harvesting apples and cleaning the river, and allow agents cleaning the river a chance to consume apples. Harvest is a public pool resource game where players must cooperate to harvest the most apples while being careful not to \qq{kill} the trees by gathering all the apples they contain. If harvesting is done too quickly, the apples will not grow back. Therefore, players must strategically distribute their harvests in order to avoid over-harvesting in any one area and hurting the trees' health.
In both games, the agents receive $+1$ extrinsic reward when they collect an apple and are punished by $-1$ extrinsic reward when they use their punishment beam as an action to fire others. The hit agent also receives $-50$ extrinsic reward. The Cleanup and Harvest environments are open-source python codes developed by Vinitsky et.al \citep{SSDOpenSource}.
\subsubsection{Experimental setup}
For Cleanup and Harvest environments, the EICM network structure used in the KindMARL method is taken from \citet{EMuReL2023} where an actor-critic structure is used to learn the policy and value functions. Since the input of this structure is a $15 \times 15$ pixels image, a single-layer convolutional network is utilized as the feature extraction network. To learn the parameters of the actor-critic structure, Proximal Policy Optimization (PPO) \citep{schulmanwdrk2017ppo} and the Asynchronous Advantage Actor-Critic (A3C) \citep{mnih2016asynchronous} algorithms are used in Cleanup and Harvest environments, respectively.
A Linux server with $3$ CPUs, a $P100$ Pascal GPU, and $100$G RAM was used to execute $7$ (resp. $5$) experiments with random seeds in Cleanup (resp. Harvest) environment. 
The number of agents operating in each run was $5$ (resp. $4$) in Cleanup (resp. Harvest) environment and they were the advantageous IA types with $\alpha = 0.05$ and $\beta=0$ for both IA and KindMARL methods.

We compared the KindMARL method with three methods: \emph{(i)} the baseline method utilizing only the extrinsic rewards, \emph{(ii)} the IA method \citep{hughes2018inequity}, and \emph{(iii)} the Social Influence (SI) method \citep{jaques2019social} that added the intrinsic rewards derived from social empowerment to the extrinsic rewards. 
\subsubsection{Experimental result}
\begin{center}
	\begin{figure}[!h]
		\centering			
		\resizebox{\columnwidth}{!}{
			\extrarowsep=_3pt^3pt			
			\begin{tabu}to\linewidth{ccc}
				\raisebox{-0.5\height}{\includegraphics[width=0.5\columnwidth ]{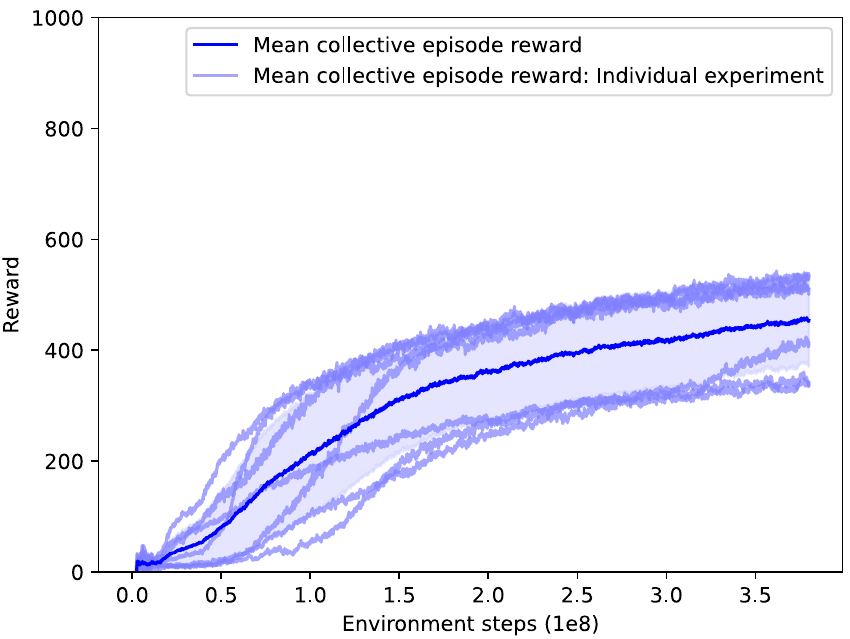}} &
				\raisebox{-0.5\height}{\includegraphics[width=0.5\columnwidth ]{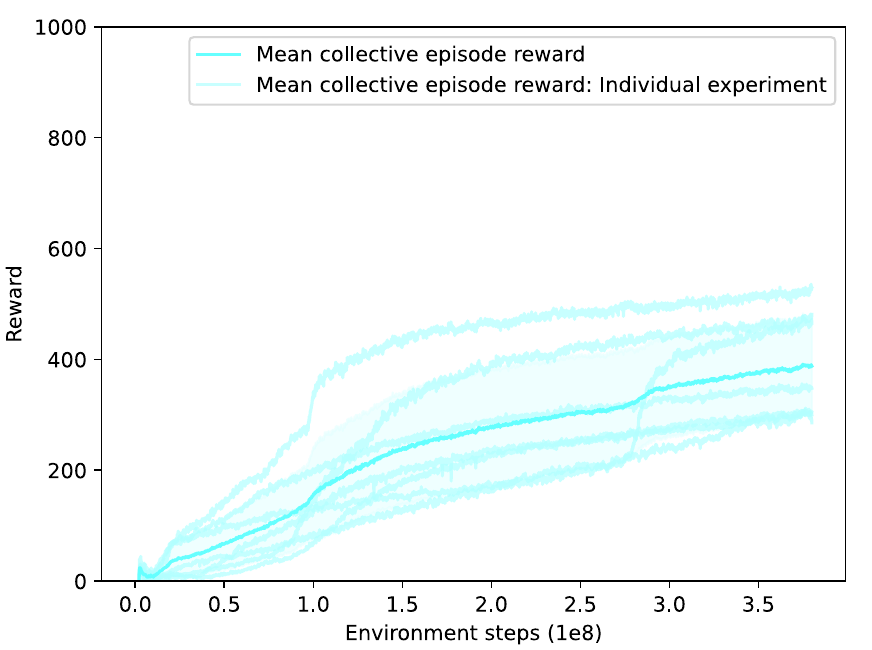}} &  
				\multirow{3}{*}{\raisebox{1.5\height}{\includegraphics[width=\columnwidth ]{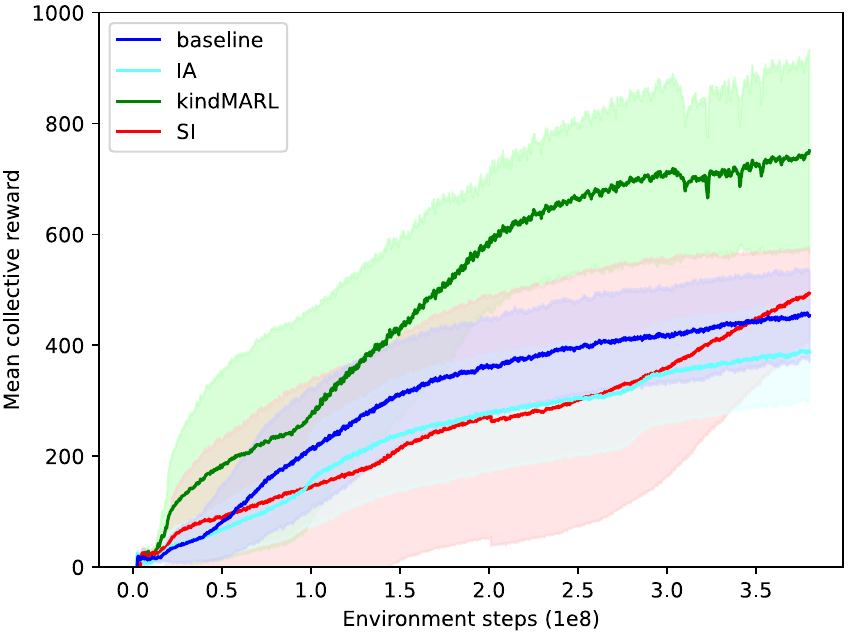}}}
				\\
				\LARGE (a) Baseline & 
				\LARGE (b) IA &  \\				
				\raisebox{-0.5\height}{\includegraphics[width=0.5\columnwidth ]{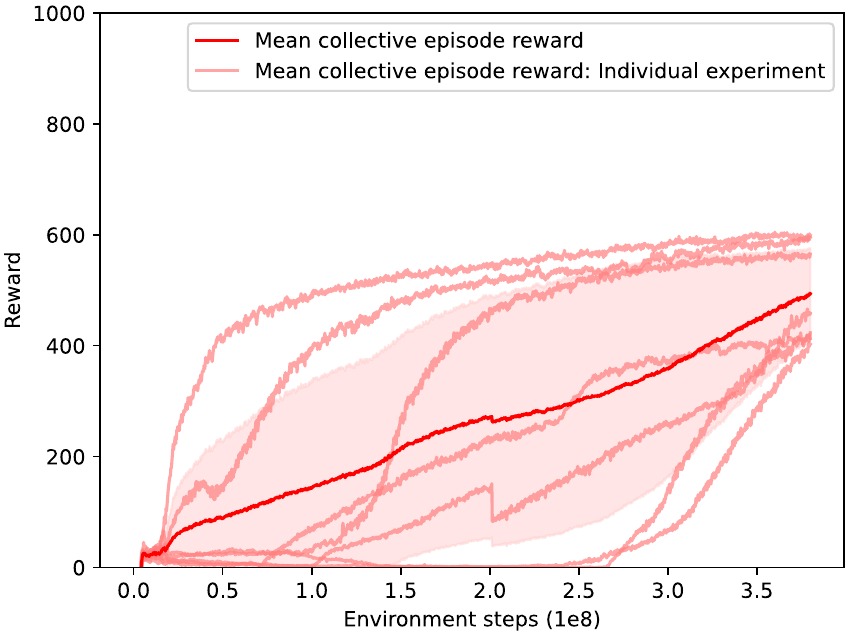}}	& \raisebox{-0.5\height}{\includegraphics[width=0.5\columnwidth ]{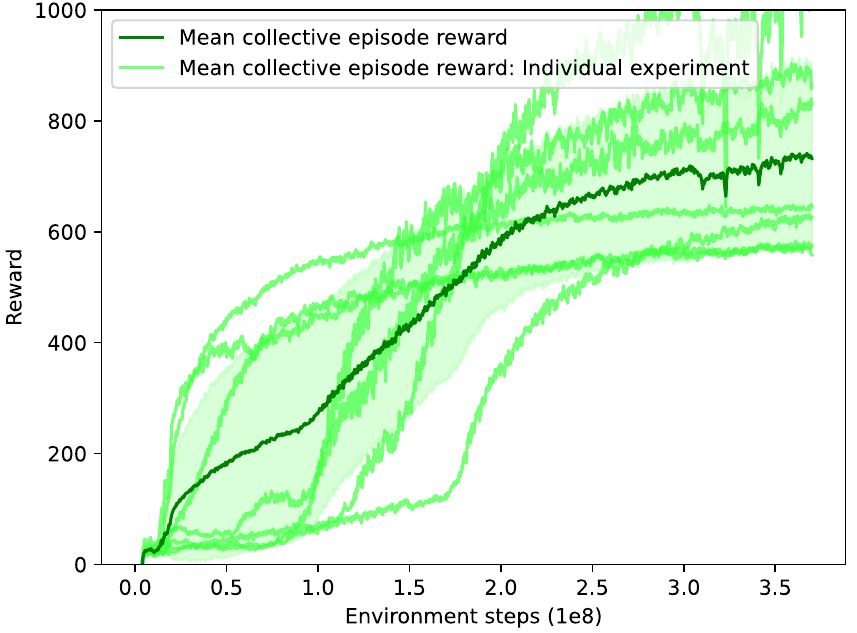}} &  \\
				\LARGE (c) SI	 & \LARGE (d)  KindMARL  & \LARGE (e) Comparison	
		\end{tabu}}
		\caption{\textbf{The results for the Cleanup environment using PPO algorithm.} The opaque curve is the mean of the results of $7$ experiments.} 
		\label{fig.Reward_cleanup_comparison}
	\end{figure}
\end{center}

\begin{center}
	\begin{figure}[!h]
		\centering			
		\resizebox{\columnwidth}{!}{
			\extrarowsep=_3pt^3pt			
			\begin{tabu}to\linewidth{ccc}
				\raisebox{-0.5\height}{\includegraphics[width=0.5\columnwidth ]{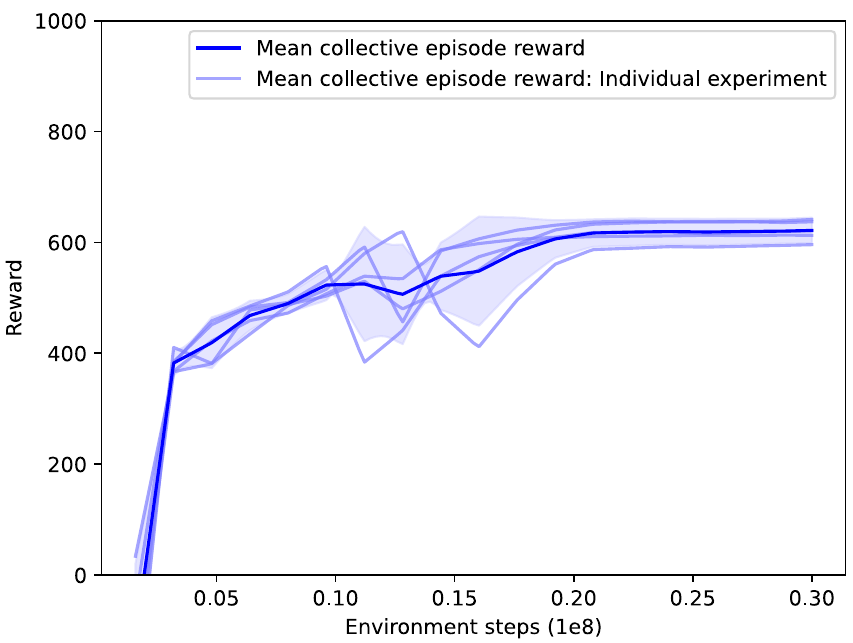}} &
				\raisebox{-0.5\height}{\includegraphics[width=0.5\columnwidth ]{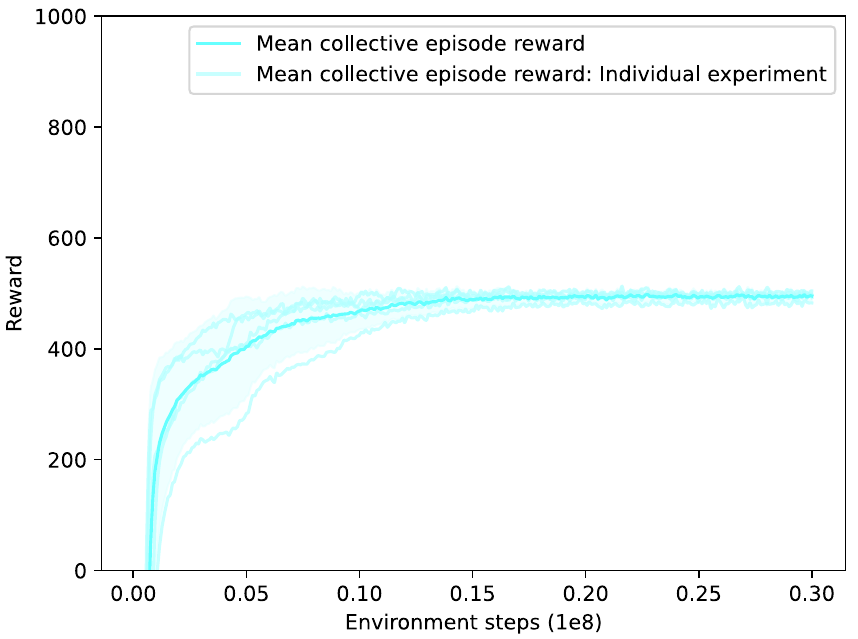}} &  
				\multirow{3}{*}{\raisebox{1.5\height}{\includegraphics[width=\columnwidth ]{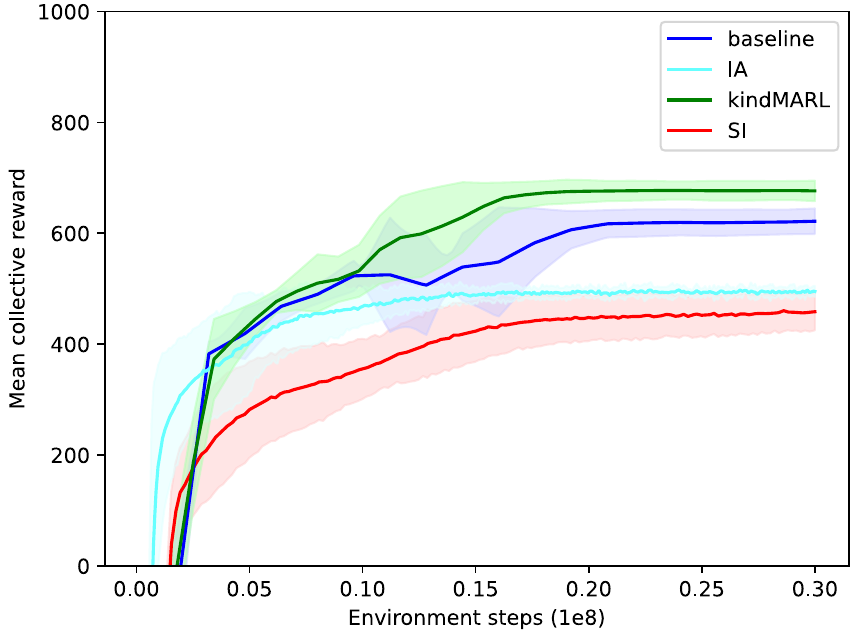}}}
				\\
				\LARGE (a) Baseline & 
				\LARGE (b) IA &  \\				
				\raisebox{-0.5\height}{\includegraphics[width=0.5\columnwidth ]{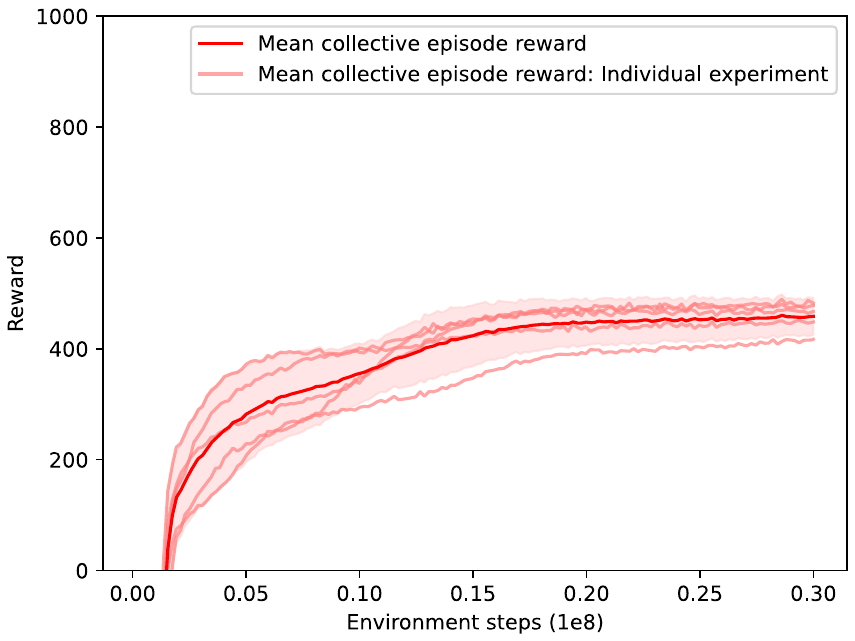}}	& \raisebox{-0.5\height}{\includegraphics[width=0.5\columnwidth ]{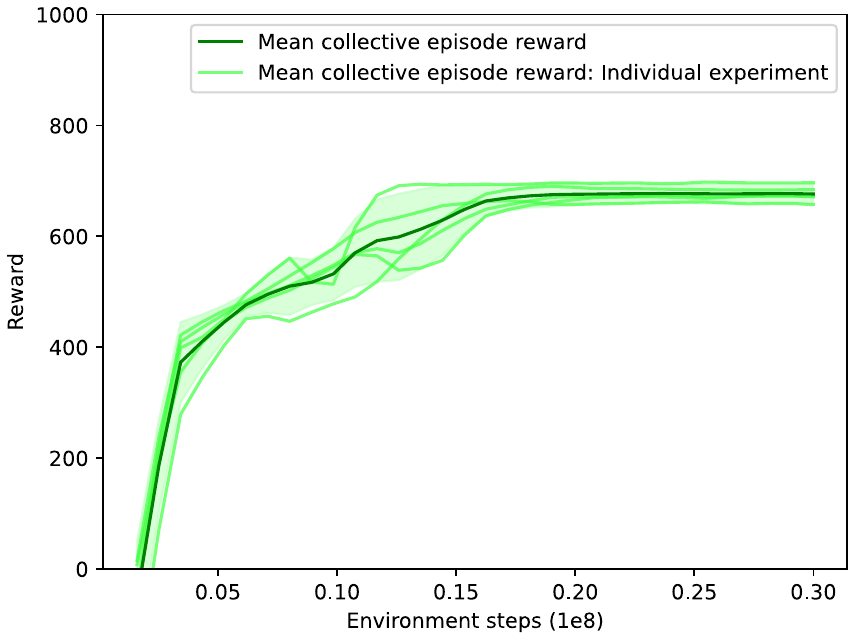}} &  \\
				\LARGE (c) SI	 & \LARGE (d)  KindMARL  & \LARGE (e) Comparison	 
		\end{tabu}}
		\caption{\textbf{The results for the Harvest environment using A3C algorithm.}  The opaque curve is the mean of the results of $5$ experiments.} 
		\label{fig.Reward_harvest_comparison}
	\end{figure}
\end{center}

We utilized The total reward received by all agents as a performance measure. According to numerical comparisons (Table \ref{Table.cleanup_harvest_expriment_results}), in Cleanup environment, training based on the KindMARL method enabled the agents to earn $60.8\%$, $88.6\%$ and $44.2\%$ more total rewards than training based on the baseline, IA model, and SI method, respectively. The KindMARL agents could also earn $9.1\%$, $37.2\%$ and $42.7\%$ more total rewards compared to baseline, IA, and SI method in Harvest environment.
The superiority of the KindMARL method is obvious in Figure. \ref{fig.Reward_cleanup_comparison} and Figure. \ref{fig.Reward_harvest_comparison} for the Cleanup and Harvest environments, respectively.
\begin{center}	
	\begin{table}[!h]
		\centering
		\caption{\textbf{Collected rewards in Cleanup and Harvest environments.} The collective reward obtained by agents in the last $\num{192000}$ and $\num{1600000}$ steps is averaged over $7$ and $5$ experiments in Cleanup and Harvest environments, respectively. The results of IA and KindMARL methods are reported for the advantageous-IA agents.}
		\label{Table.cleanup_harvest_expriment_results}
		\resizebox{0.5\columnwidth}{!}{
			\extrarowsep=_3pt^3pt
			\begin{tabu}to\linewidth{|c|c|[2pt gray]c|c|}				
				\cline{3-4}
				\multicolumn{2}{c|[2pt gray]}{} & \multicolumn{2}{c|}{\textbf{Environment (Algorithm)}} \\	
				\cline{3-4}
				\multicolumn{2}{c|[2pt gray]}{} & Cleanup (PPO)  & Harvest (A3C) \\				
				\cline{2-4}				
				\tabucline[1.5pt gray]-
				\multirow{4}{*}{\textbf{Methods}} & Baseline & 470.9 & 622.2  \\
				\cline{2-4}
				& IA & 401.6 & 494.7 \\	
				\cline{2-4}
				& SI & 525.3 & 475.5 \\				
				\cline{2-4}
				& KindMARL & \textbf{757.4} & \textbf{678.5} \\				
				\cline{1-4}
		\end{tabu}}
	\end{table}
\end{center}
\subsection{The problem of traffic light control}
\label{S.Experiments.Environment}
We consider the problem of traffic light control to cooperate between intersections, the traffic generating situations where a large number of vehicles have to wait for traffic permits to reach their destinations while avoiding accidents. Many researchers took into account this challenge and introduced various methods of controlling traffic signals to improve transportation efficiency. 

Besides many conventional methods, the problem was recently tackled with RL methods. 
Some examples are as follows: \emph{(i)}  IntelliLight \citep{wei2018intellilight} used real data collected from surveillance cameras and trained a DQN method to control the traffic lights of one intersection in SUMO simulator \citep{SUMO2018}. It evaluated the learned policy not only by obtained rewards but also by considering the quality of the pattern learned for light switching.
\emph{(ii)} DemoLight \citep{xiong2019learning} utilized the Self-Organizing Traffic Light Control \citep{cools2013self} to make the agents learn from an expert using CityFlow simulator, that is an open-source traffic simulator designed for large-scale traffic scenarios \citep{zhang2019cityflow}. To speed up the learning process, DemoLight applied the Advantage Actor-Critic (A2C) algorithm \citep{mnih2016asynchronous}. 
\emph{(iii)} PressLight \citep{wei2019presslight} defined the reward function according to Max Pressure method \citep{lioris2016adaptive} and trained a DQN structure to control the traffic lights of several intersections in CityFlow simulator. 
\emph{(iv)} MetaLight \citep{zang2020metalight}  applied the gradient-based meta-learning algorithm \citep{finn2017model} in CityFlow simulator to speed up the learning process and improve generalization so that the knowledge gained in previous traffic scenarios can be used in new ones. This method also improved the Flipping and Rotation and considers All Phase (FRAP) model \citep{zheng2019learning}, that is based on the DQN structure to train the agents.
\emph{(v)} MPLight method \citep{chen2020toward} is another DQN based method that employed a shared DQN structure to control all intersections and also used the concept of pressure in traffic. It conducted the experiments on an environment with more than $1,000$ traffic lights using CityFlow simulator. 
\emph{(vi)} Colight \citep{wei2019colight} utilized the graph attention network \citep{velivckovic2017graph}, that represents the neighboring intersections as an overall summary and enables the agents to learn the model of neighboring influence on their under-control intersection. Each agent managed an intersection in CityFlow simulator and had an instance of a DQN structure.

\subsubsection{Experimental setup}
\label{S.Experiments.Experimental_Setup}
We utilized the Colight method \citep{wei2019colight} as a baseline to implement the KindMARL method. Colight creates some environments on the CityFlow simulator to implement traffic light signal control methods by collecting data from Chinese cities such as Hangzhou and Jinan and New York City in the United States, as well as artificially generated data. In this environment, several vehicles move from various origins to their destinations while encountering traffic lights. The green signal of each traffic light is accompanied by a yellow three-second signal and a two-second red signal.

We conducted our experiments on the environment that has been created by using Hangzhou dataset, which consists of a $4\times4$ grid ($16$ intersections). Each agent controls an intersection and sends its information to its four neighbors that located at its four main directions. To learn appropriate polices in this environment, each experiment includes several episodes. Each episode contains $1440$ samples, that is the maximum amount of data that can be created in an intersection during a day. 

Here in EICM Structure, each agent learns the parameters of a Deep Q-Network (DQN) structure \citep{mnih2015human} as $Q^{\pi^k}(s^k_t,a^k_t)$, that receives the visited local state $s^k_t$ as input and estimates the \textit{state-action value} of applying each available action of agent $k$ in the local state $s^k_t$ as output (Figure. \ref{fig.EICM}).
\begin{center}
	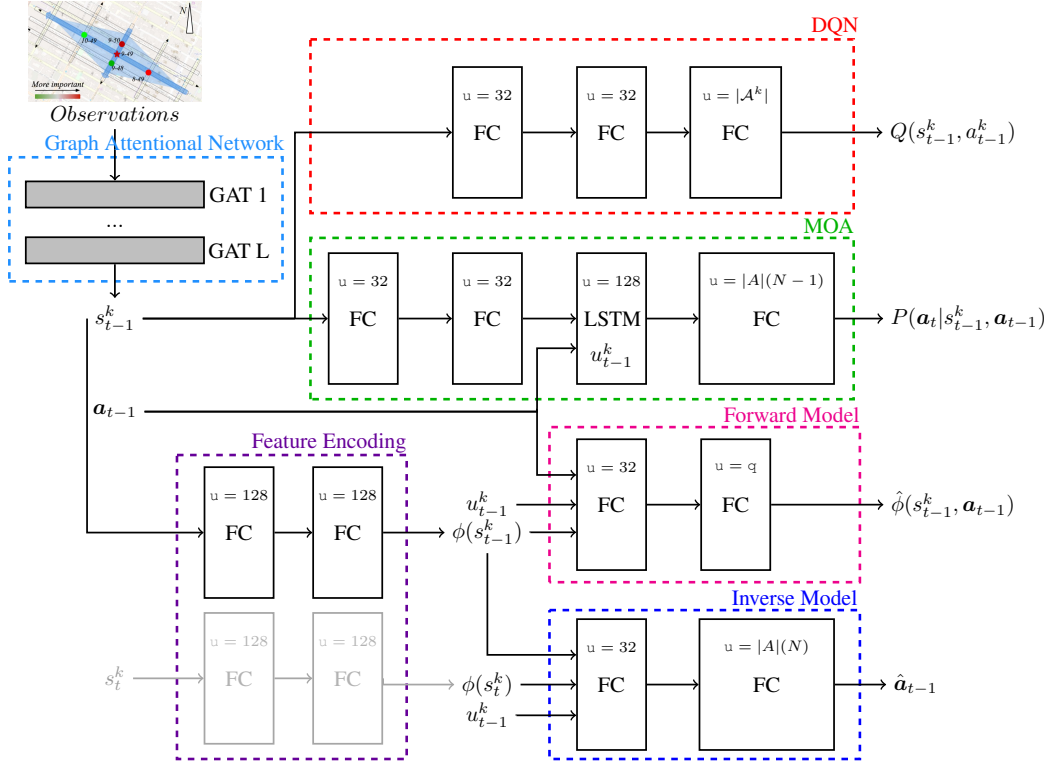
\begin{figure}[!h]	
		\centering
		\resizebox{\columnwidth}{!}{
			\begin{tikzpicture}
			\node[inner sep=0pt] (image) at (0,4.3)
			{\includegraphics[width=.2\textwidth]{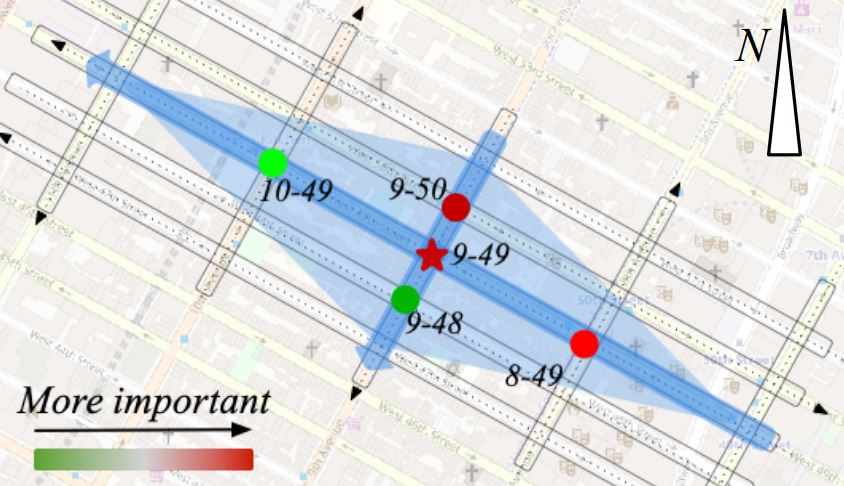}};			
			\node (obs) at (0,3.3) [] {$Observations$};
			\node (GAT1) at (0,2) [layers, rectangle, text width=7.5em, minimum height=1.2em, fill=gray!50!white] {};
			\node (GAT1Text) at (2,2) [] {GAT 1};
			\node (points) at (0,1.5) [] {$...$};			
			\node (GATL) at (0,1.1) [layers, rectangle, text width=7.5em, minimum height=1.2em, fill=gray!50!white] {};
			\node (GATLText) at (2,1.1) [] {GAT L};
			\node (Input1) at (0,0) [] {$s^k_{t-1}$};
			\draw [->, thick] (obs.south) -- (GAT1.north);  
			\draw [->, thick] (GATL.south) -- (Input1.north); 			
			\node (GAN) at (0.8,2.8) [text=dodgerblue ] {Graph Attentional Network}; 
			
			\node (Input2) at (0,-1.5) [] {$\bm{a}_{t-1}$};						
			\node (FC1)  at (4,0) [layers, rectangle] {FC};
			\node (FC1u) at (4,0.6) [] {\scriptsize $\texttt{u}=32$};
			\node (FC2)  at (6,0) [layers, rectangle] {FC};
			\node (FC2u) at (6,0.6) [] {\scriptsize $\texttt{u}=32$};
			\node (LSTM) at (8,0) [layers, rectangle] {LSTM};
			\node (LSTMu) at (8,0.6) [] {\scriptsize $\texttt{u}=128$};
			\node (ukt) at (8,-0.6) [] {$u^k_{t-1}$};
			\node (FC3)  at (10.5,0) [layers, rectangle, text width=5.5em] {FC};
			\node (FC3u) at (10.5,0.6) [font=\scriptsize] { $\texttt{u}=|A|(N-1)$};
			\node (Output) at (13.75,0) [] {$P(\bm{a}_{t}|s^k_{t-1},\bm{a}_{t-1})$ };		 
			\draw [->, thick] (Input1.east) --  (FC1.west);
			\draw [->, thick] (FC1.east) --  (FC2.west);
			\draw [-, thick] (Input2) -| (6.8,-1.45) ;
			\draw [->, thick] (6.8,-1.45) |- (LSTM.220) ;
			\draw [->, thick] (FC2.east) --  (LSTM.west);    
			\draw [->, thick] (LSTM.east) --  (FC3.west);    
			\draw [->, thick] (FC3.east) -- (Output.west);      
			\node (MOA) at (11.5,1.5) [text=black!30!green] {MOA}; 
			
			\node (dqFC1)  at (6,3) [layers, rectangle] {FC};
			\node (dqFC1u) at (6,3.6) [] {\scriptsize $\texttt{u}=32$};			
			\node (dqFC2) at (8,3) [layers, rectangle] {FC};
			\node (dqFC2u) at (8,3.6) [] {\scriptsize $\texttt{u}=32$};		
			\node (dqFC3q)  at (10,3) [layers, rectangle, text width=3.5em] {FC};			
			\node (dqFC3qu) at (10,3.6) [font=\scriptsize] { $\texttt{u}=|\mathcal{A}^k|$};			
			\node (dqOutputq) at (13.5,3) [] {$Q(s^k_{t-1},a^k_{t-1})$};		
			\draw [-, thick] (Input1.east) --  (2.9,0);
			\draw [->, thick] (2.9,0) |-  (dqFC1.west);
			\draw [->, thick] (dqFC1.east) --  (dqFC2.west);
			\draw [->, thick] (dqFC2.east) --  (dqFC3q.west);    					
			\draw [->, thick] (dqFC3q.east) -- (dqOutputq.west);
			\node (DQN) at (11.6,4.7) [text=red] {DQN};
			
			\node (FAFC1) at (2,-3.45) [layers, rectangle, text width=2.5em] {FC};
			\node (FAFC1u) at (2,-2.85) [] {\scriptsize $\texttt{u}=128$};
			\node (FAFC2) at (3.75,-3.45) [layers, rectangle, text width=2.5em] {FC};
			\node (FAFC2u) at (3.75,-2.85) [] {\scriptsize $\texttt{u}=128$};	
			\draw [->, thick] (Input1.west) |- (FAFC1.west);
			\draw [->, thick] (FAFC1.east) |- (FAFC2.west);
			\node (PhiPrevSInput) at (6,-3.45) [] {$\phi(s^k_{t-1})$};
			\draw [->, thick] (FAFC2.east) |- (PhiPrevSInput.west);
			\node (SA2Input) at (0,-5.8) [text=gray!80] {$s^k_{t}$};
			\node (FA2FC1) at (2,-5.8) [layers, rectangle, draw=gray!70, text=gray!60, text width=2.5em] {FC};
			\node (FA2FC1u) at (2,-5.2) [text=gray!60] {\scriptsize $\texttt{u}= 128$};		
			\node (FA2FC2) at (3.75,-5.8) [layers, rectangle, draw=gray!70, text=gray!60, text width=2.5em] {FC};
			\node (FA2FC2u) at (3.75,-5.2) [text=gray!60] {\scriptsize $\texttt{u}=128$};			
			\draw [->, thick, draw=gray!70] (SA2Input.east) -- (FA2FC1.west);
			\draw [->, thick, draw=gray!70] (FA2FC1.east) -- (FA2FC2.west);
			\node (PhiSInput) at (6,-5.9) [] {$\phi(s^k_{t})$};
			\draw [->, thick, draw=gray!70] (FA2FC2.east) |- (PhiSInput.west);
			\node (FeatureEncoding) at (3.45,-2) [text=blue!60!red] {Feature Encoding};
			
			\node (uktInputF) at (6,-3) [] {$u^k_{t-1}$};
			\node (FFC1)  at (8,-3.0) [layers, rectangle] {FC};
			\node (FFC1u) at (8,-2.4) [] {\scriptsize $\texttt{u}=32$};
			\node (FFC2)  at (10,-3.0) [layers, rectangle] {FC};
			\node (FFC2u) at (10,-2.4) [] {\scriptsize $\texttt{u}=\texttt{q}$};
			\node (FOutput) at (13.5,-3.0) [] {$\hat{\phi}(s^k_{t-1},\bm{a}_{t-1})$};
			\draw [-, thick] (Input2) -| (6.8,-1.45) ;
			\draw [->, thick] (6.8,-1.45) |- (FFC1.140);
			\draw [->, thick] (uktInputF.east) |- (FFC1.180);
			\draw [->, thick] (PhiPrevSInput.east) |- (FFC1.218);
			\draw [->, thick] (FFC1.east) |- (FFC2.west);
			\draw [->, thick] (FFC2.east) |- (FOutput.west);
			\node (ForwardModel) at (10.9,-1.55) [text=magenta] {Forward Model};
			
			\node (uktInputI) at (6,-6.4) [] {$u^k_{t-1}$};
			\node (IFC1)  at (8,-5.9) [layers, rectangle] {FC};
			\node (IFC1u) at (8,-5.3) [] {\scriptsize $\texttt{u}=32$};
			\node (IFC2)  at (10.5,-5.9) [layers, rectangle, text width=5.5em] {FC};
			\node (IFC2u) at (10.5,-5.3) [] {\scriptsize $\texttt{u}=|A|(N)$};
			\node (IOutput) at (12.9,-5.9) [] {$\hat{\bm{a}}_{t-1}$};
			\draw [->, thick] (PhiPrevSInput.south) |- (IFC1.140);
			\draw [->, thick] (PhiSInput.east) |- (IFC1.180);
			\draw [->, thick] (uktInputI.east) |- (IFC1.221);
			\draw [->, thick] (IFC1.east) |- (IFC2.west);
			\draw [->, thick] (IFC2.east) |- (IOutput.west);
			\node (InverseModel) at (10.95,-4.5) [text=blue] {Inverse Model};
			
			\begin{pgfonlayer}{background}  
			
			\path[draw=red, dashed, very thick] (3.15,4.5)--(11.9,4.5)-- (11.9,1.7)--(3.15,1.7)--(3.15,4.5);
			
			\path[draw=black!30!green, dashed, very thick]
			(3.15,-1.3) rectangle (11.9,1.3); 
			
			\path[draw=blue!60!red, dashed, very thick] (1,-2.2)-- (4.7,-2.2)--(4.7,-7.1)-- (1,-7.1)--(1,-2.2); 
			
			\path[draw=magenta, dashed, very thick] (7,-4.25)-- (7,-1.75)--(12,-1.75)-- (12,-4.25)--(7,-4.25); 
			
			\path[draw=blue, dashed, very thick] (7,-4.7)-- (12,-4.7)-- (12,-7.1)--(7,-7.1)--(7,-4.7); 
			
			\path[draw=dodgerblue, dashed, very thick] (-1.7,0.6)-- (2.7,0.6)-- (2.7,2.6)--(-1.7,2.6)--(-1.7,0.6);

			\end{pgfonlayer}
			\begin{pgfonlayer}{foreground} 
			
			\end{pgfonlayer}		
			\end{tikzpicture}}
		\caption{\textbf{The EICM network structure used in the Traffic light control problem.} The inputs of each agent $k$'s EICM are the previous joint action $\bm{a}_{t-1}$ and the previous local state $s^k_{t-1}$ that is composed of the representing features provided by the output of the graph attentional network that is utilized for graph-structured environment where a summary of each graph node's neighborhood is prepared by embedding information from the neighbors and their importance to the central node. The $L$ graph attention layers of this network are illustrated by GAT. The EICM uses the DQN structure to learn the state-action value function, the MOA network to predict the next action of other agents, the feature encoding network to extract \texttt{q} features from the local state $s^k_{t-1}$, the forward model to learn $\hat\phi$, and the inverse model to predict the applied actions. In each time step $t$, feature encoding is done for the local states $s^k_{t-1}$ and $s^k_t$. The layers represented by the transparent gray rectangles inside the purple dashed rectangle just indicates the same operation done for $s^k_t$. Fully connected layers are indicated by FC, and the parameter \texttt{u} determines the number of their neurons. To predict each agent $j$'s action $a^{j}_{t}$, agent $k$'s MOA captures the state transition in its internal LTSM state $u^k_{t-1}$. The graph attentional network and DQN structures are taken from \citet{wei2019colight}. The MOA structure is taken from \citet{jaques2019social}. The forward and inverse model are based on the structures presented by \citet{heemskerk2020social,EMuReL2023}.} 
		\label{fig.EICM}
	\end{figure}
\end{center}

We compared the KindMARL method with two methods: \emph{(i)} the CoLight method \citep{wei2019colight}, and \emph{(ii)} the IA method \citep{hughes2018inequity}. 
Since the goal of the traffic light control problem is minimizing the vehicles' travel times, the extrinsic reward received by each agent is defined by the quarter of the collective time of all vehicles stopped at its associated intersection. 
To evaluate the performance of each method, the time distance between the entering and leaving the grid for each vehicle was measured and their average was used as the vehicle travel time metric. 
We executed each method $3$ times and took an average between the obtained evaluation metric.	
For the CoLight method, we have set the number of the heads in the attention mechanism to $5$. 
For the IA, and KindMARL methods, first we used the typical advantageous type of IA model where the $\alpha$ is zero and $\beta$ is set to $0.05$ and executed experiments including a length of $200$ episodes. Then, we searched the range of $-1$ to $1$ for the $\alpha$ and $\beta$ hyperparameters in the IA model to investigate the effect of agents' aversion to each of the advantageous and disadvantageous inequities and to find the best combination of these coefficients. To reduce the execution time, this search was done for IA and KindMARL methods separately by executing experiments including a length of $100$ episodes.
Each experiment took around $8$ hours to perform on a Linux server with $16$ CPUs and $120$G RAM.				
\subsubsection{Experimental results}
\label{S.Experiments.Experimental_Results}
\begin{center}
	\begin{figure}[!h]
		\centering			
		\resizebox{\columnwidth}{!}{
			\extrarowsep=_3pt^3pt			
			\begin{tabu}to\linewidth{c}
				{\includegraphics[width=0.5\columnwidth ]{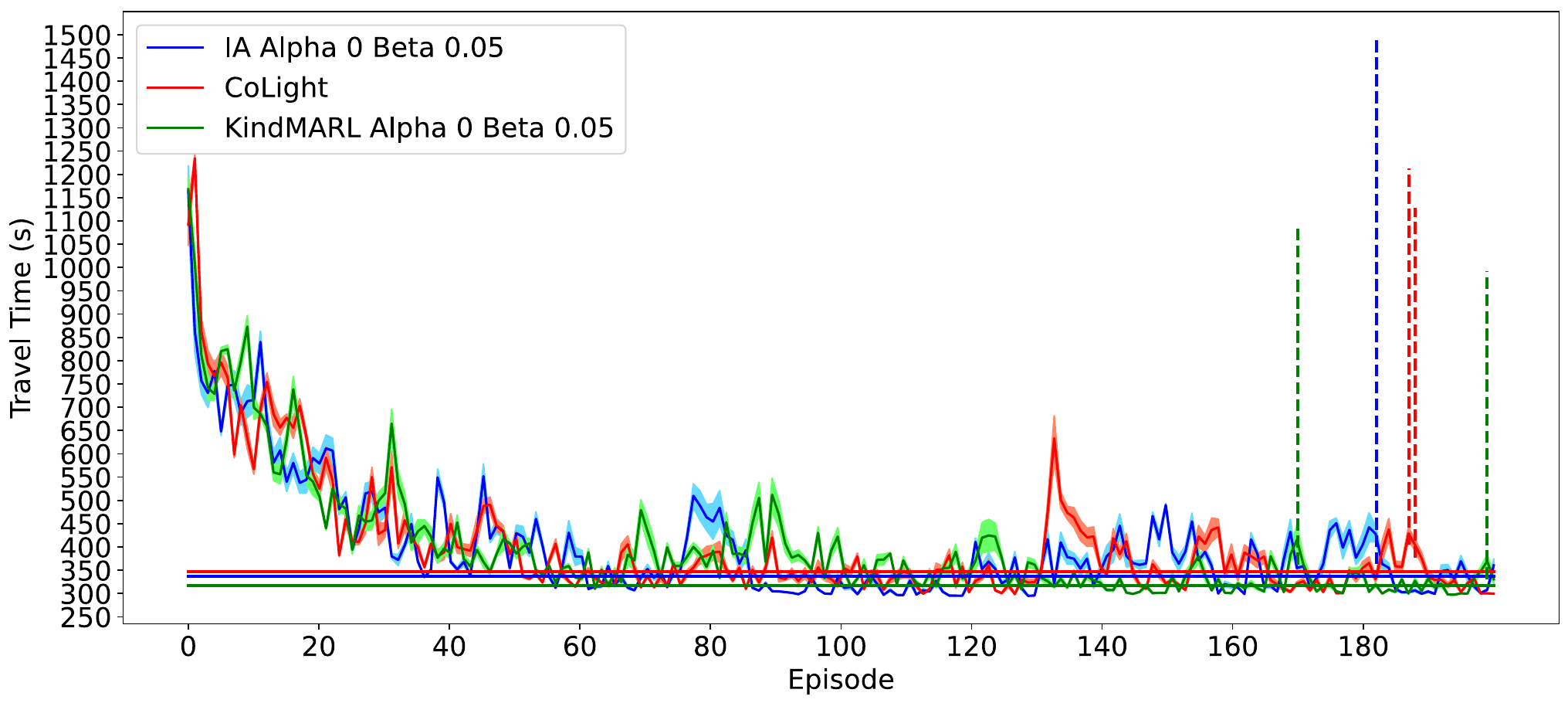}}  
		\end{tabu}}
		\caption{\textbf{The comparison between the mean travel times of all methods using the typical advantageous IA agents in the IA and KindMARL methods.} The reported delay time at each point of these curves is computed by considering all $16$ agents and over $3$ individual experiments. The vertical dashed lines illustrated with same color as each method's curve represent the position where the method converges. The closer this line is to the vertical axis, the faster is the method's convergence. To draw these lines, we calculated the average travel time obtained in the last $20$ episodes of each method as a threshold. Then we checked if there is an episode in range of $50$ and $200$ that the maximum travel time computed for each episode from that episode to end is less that $1.1$ and $1.2$ times of the threshold. If the episodes with the first and second conditions were found, we determined the first and second vertical lines at those episode numbers, respectively. The horizontal lines indicate the average of the last $20$ steps of each episode, that the lower line means the less travel time of the method. The result of the KindMARL method is the best considering the above evaluation metrics.}
		\label{fig.Reward_Comparison}
	\end{figure}
\end{center}

\begin{center}
	\begin{figure}[!h]
		\centering			
		\resizebox{\columnwidth}{!}{
			\extrarowsep=_3pt^3pt			
			\begin{tabu}to\linewidth{cc}
				{\includegraphics[width=0.5\columnwidth ]{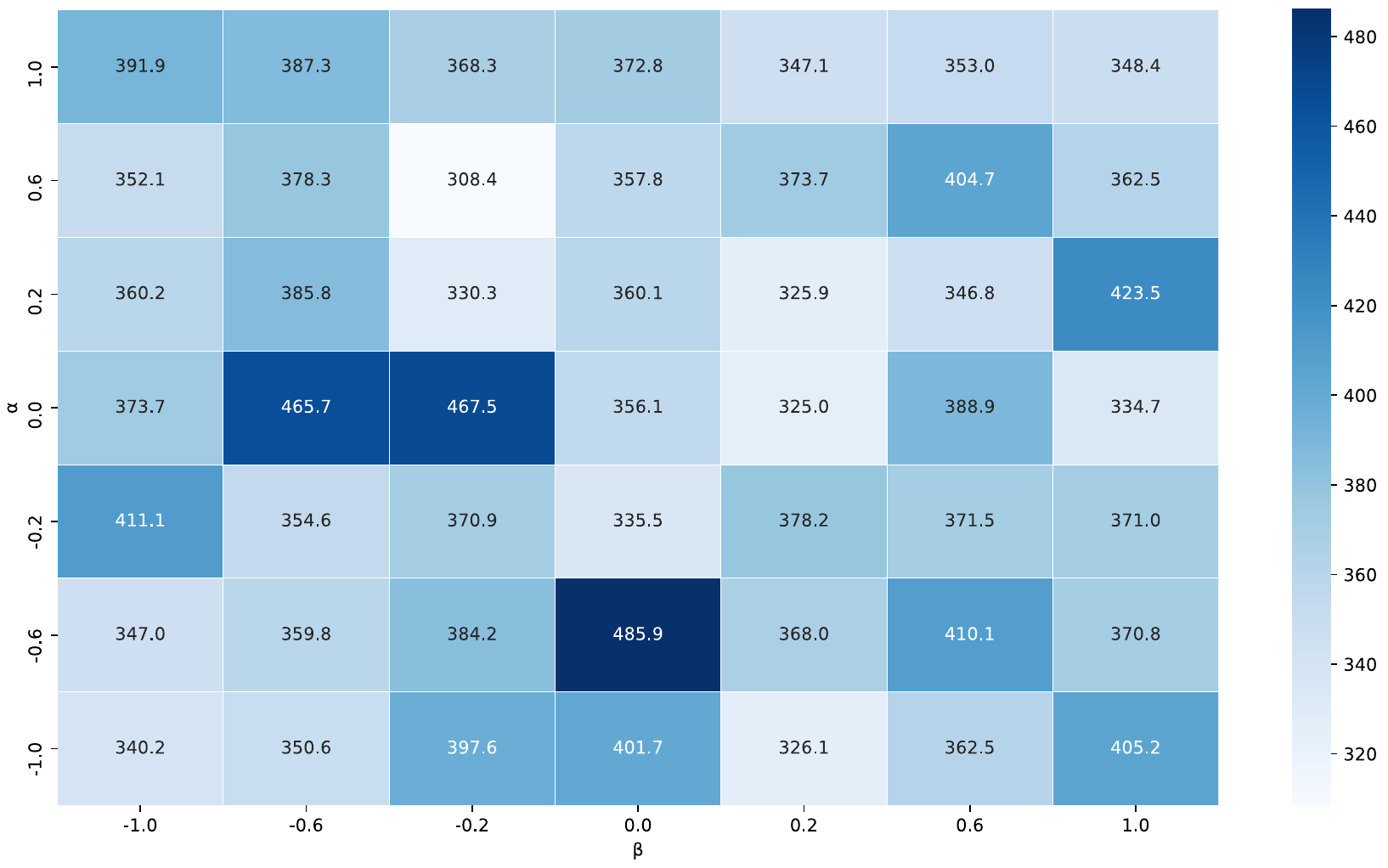}} &
				{\includegraphics[width=0.5\columnwidth ]{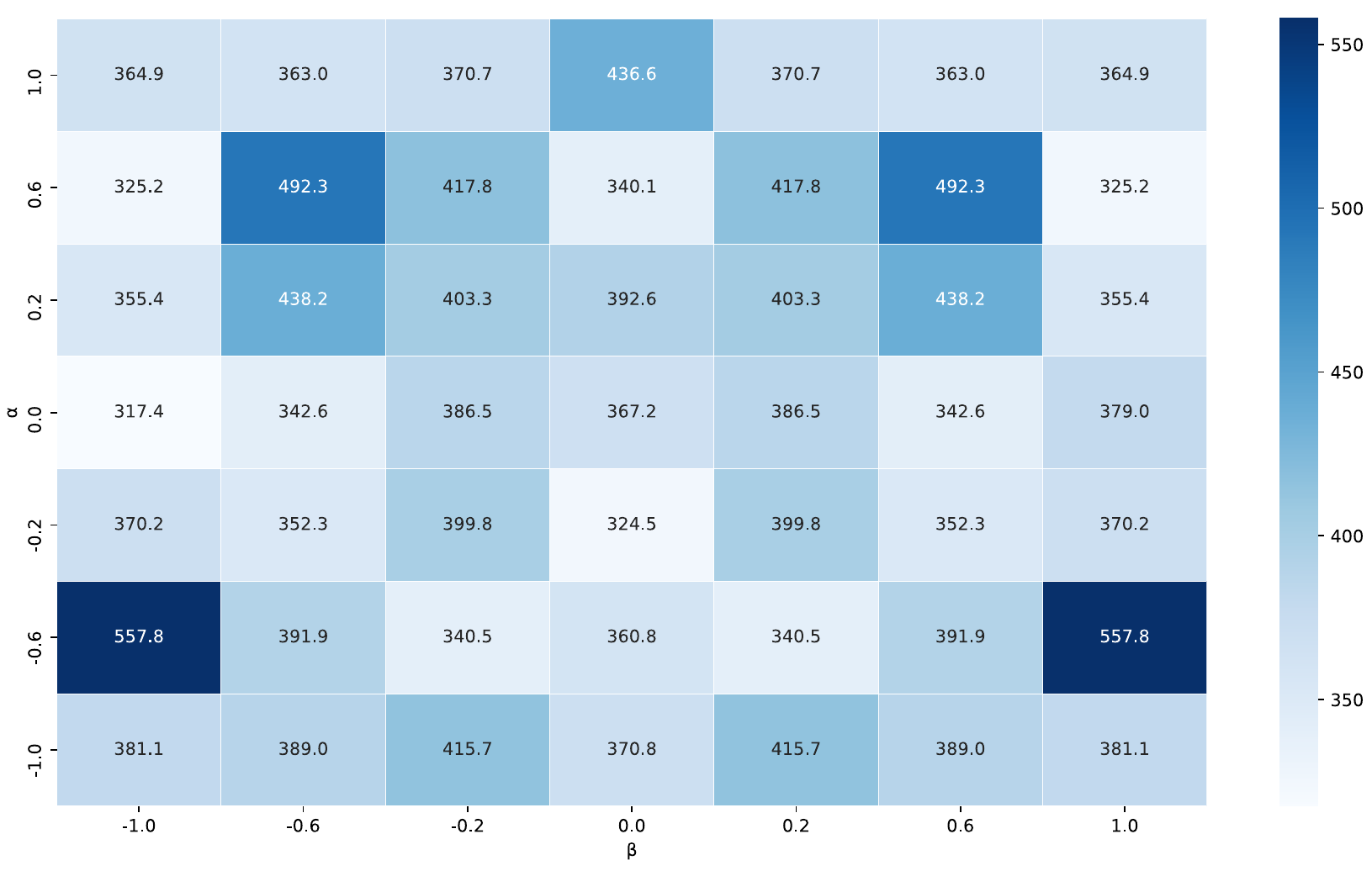}} \\
				(a) & (b)
		\end{tabu}}
		\caption*{\textbf{The results of the average travel time over the last $\textbf{20}$ episodes achieved by the $\textbf{3}$ experiments done for each IA model's $\bm{\alpha}$ and $\bm{\beta}$ combination using the Hangzhou dataset.} (a) The search result for the IA method, (b) The search result for the KindMARL method. In each figure, the result of CoLight is obtained by $\alpha=0, \beta=0$.}
		\label{fig.IA_Kind_AlphaBetaSearch}
	\end{figure}
\end{center}

\begin{center}
	\begin{figure}[!h]
		\centering			
		\resizebox{\columnwidth}{!}{
			\extrarowsep=_3pt^3pt			
			\begin{tabu}to\linewidth{c}			
				{\includegraphics[width=\columnwidth ]{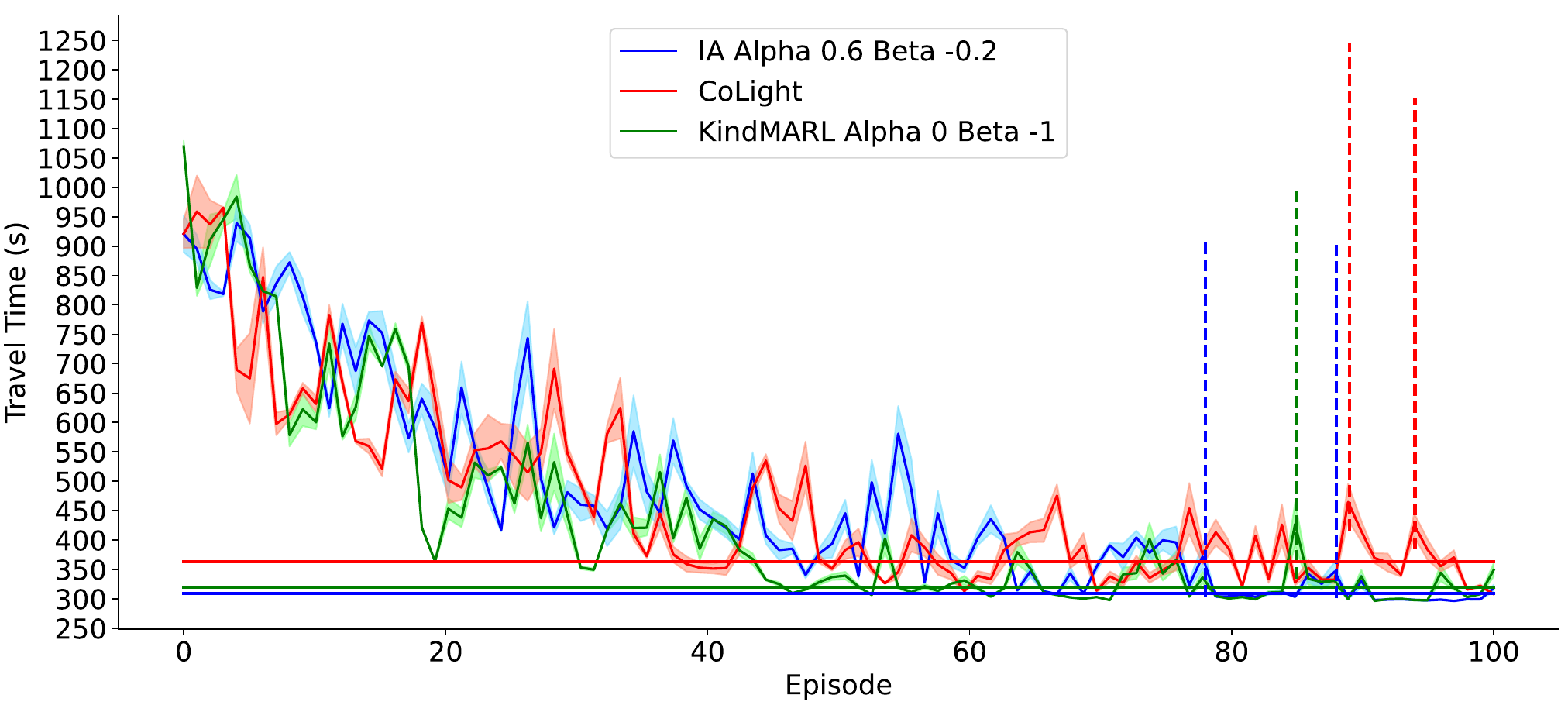}}
				
		\end{tabu}}
		\caption{\textbf{The comparison between the mean travel times of all methods using the best combination of IA model's hyperparameters in the IA and KindMARL methods.} To draw vertical dashed lines, we did the same operation as described in Figure. \ref{fig.Reward_Comparison} except reducing the range of episodes from $200$ to $100$. The result of CoLight is obtained by $\alpha=0, \beta=0$.}
		\label{fig.AlphaBetaSearch_comarison}
	\end{figure}
\end{center}

According to the results obtained by the typical advantageous IA agents in IA and KindMARL methods (Figure \ref{fig.Reward_Comparison}), the convergence speed of the KindMARL method is better than others, but the IA and CoLight methods have a lower distance between the two convergence indexes. The average travel time computed for the last $20$ episodes demonstrates that the KindMARL outperforms other methods. 
The exact values of the average travel times reported for the methods also confirm the superiority of the KindMARL method (Table. \ref{Table.experiment_results}). It is observed that the KindMARL method had the best average travel time of the last $20$ steps of each episode as the final learned performance and it is better than the CoLight and IA methods by $8.8\%$, and $5.9\%$, respectively.	
The $\alpha$ and $\beta$ search for the IA and KindMARL methods (Figure. \ref{fig.IA_Kind_AlphaBetaSearch}) resulted in some combinations of these parameters that improve the travel time. The results of the IA and KindMARL methods with the best $\alpha$ and $\beta$ combination (Figure. \ref{fig.AlphaBetaSearch_comarison}, Table. \ref{Table.Search_experiment_results}) also demonstrate the IA and KindMARL methods obtain lower travel times and better convergence speed than the CoLight method.
\begin{table}[h]
	\begin{center}
		\begin{minipage}{\textwidth}
			\caption{\textbf{Statistical results for the Hangzhou dataset using the typical advantageous IA agents in the IA and KindMARL methods.}}
			\label{Table.experiment_results}
			\begin{tabular*}{\textwidth}{@{\extracolsep{\fill}}lccc@{\extracolsep{\fill}}}
				\toprule				
				Methods & Episode & Performance & Convergence  \\
				& (From $20$ to $200$) & (Second) & (Episode numbers)  \\
				\toprule
				CoLight  & $366.8$ & $347.4$ & $ 187, 188$  \\
				\midrule				
				IA &   &   &   \\
				$\alpha = 0 , \beta = 0.05$ & $373.4$ & $336.7$  & $182, 182$   \\
				\midrule				
				KindMARL &   &   &   \\
				$\alpha = 0 , \beta = 0.05 $ & $\textbf{364.9}$ & $\textbf{316.9}$  & $\textbf{170}, 199$   \\
				\toprule			
			\end{tabular*}
			\footnotetext{Note: The first numerical column displays the average travel time obtained by whole vehicles in episodes from $20$ to $200$ over $3$ individual experiments of CoLight, IA and KindMARL for the Hangzhou dataset. The second and third numerical columns indicate the average of the last $20$ steps of each episode as the final learned performance and the convergence thresholds, respectively. The $\alpha$ and $\beta$ values are reported for the advantageous type of IA model typically used in IA method.
			}					
		\end{minipage}
	\end{center}
\end{table}  		
\begin{table}[h]
	\begin{center}
		\begin{minipage}{\textwidth}
			\caption{\textbf{Statistical results for the Hangzhou dataset using the best combination of IA model's hyperparameters in the IA and KindMARL methods.}}
			\label{Table.Search_experiment_results}
			\begin{tabular*}{\textwidth}{@{\extracolsep{\fill}}lccc@{\extracolsep{\fill}}}
				\toprule				
				Methods & Episode & Performance & Convergence  \\
				& (From $20$ to $100$) & (Second) & (Episode numbers)  \\
				\toprule
				CoLight  & $407.9$ & $363.4$ & $ 89,94$  \\
				\midrule				
				IA (Exhaustive search)&   &   &   \\
				$\alpha = 0.6 , \beta = -0.2$ & $398.43$ & $\textbf{309.1}$  & $\textbf{78},\textbf{88}$   \\
				\midrule				
				KindMARL (Exhaustive search)&   &   &   \\
				$\alpha = 0 , \beta = -1$ & $\textbf{360.9}$ & $319.9$  & $85,85$   \\
				\toprule			
			\end{tabular*}
			\footnotetext{Note: The numerical columns display the same information as described in Figure. \ref{Table.experiment_results} except the first column showing the average travel time obtained by whole vehicles in episodes from $20$ to $100$ over $3$ individual experiments of CoLight, IA, and KindMARL for the Hangzhou dataset.  The $\alpha$ and $\beta$ values are reported for the best results of different types of IA model considering the positive and negative ranges of these coefficients.
			}					
		\end{minipage}
	\end{center}
\end{table}

\section{Conclusion}
\label{S.Discussion}
We introduced a method of modeling the kindness concept in human societies to improve the cooperativeness of agents trained in MARL problems. 
To determine the kindness of the other agents, each agent evaluates the intentions of its fellows. 
Then it uses the intentions as the weights of the agents' rewards comparison to improve the intrinsic reward functions of the IA method to encourage agents to cooperate effectively in the MARL environments. 
To demonstrate the superiority of the proposed KindMARL method, we conducted the experiments on real data from Hangzhou city and compared the result of the KindMARL with three state-of-the-art methods IA, EMuReL, and Colight. 
The comparison results showed that the KindMARL method reduces the average travel delay of the vehicles more than the other methods.

These experiments can be performed considering more episodes on data in larger volumes to evaluate the KindMARL method for more realistic environments. 
An exhaustive search over the IA method parameters can also improve the final results. 
These cases are postponed to future work.

\begin{ack}
We would like to thank Digital Research Alliance of Canada for providing computational resources that facilitated our experiments.
\end{ack}

\bibliographystyle{plainnat-reversed}
\bibliography{mybib}

\end{document}